%% file: visda.tex
\documentclass[runningheads]{llncs}
\input{math_commands.tex}

\usepackage{graphicx}
\usepackage{floatrow}
\usepackage{comment}
\usepackage{amsmath,amssymb} 
\usepackage{color}
\usepackage{epstopdf}
\newfloatcommand{figurebox}{figure}[\nocapbeside][\dimexpr(\textwidth-\columnsep)/2\relax]
\newfloatcommand{tablebox}{table}[\nocapbeside][\dimexpr(\textwidth-\columnsep)/2\relax]

\usepackage{diagbox}

\usepackage{array}
\newcolumntype{C}[1]{>{\centering\arraybackslash}m{#1}}
\newcolumntype{R}[1]{>{\raggedleft\arraybackslash}m{#1}}
\newcolumntype{P}[1]{>{\raggedright\arraybackslash}p{#1}}
\newcolumntype{M}[1]{>{\centering\arraybackslash}m{#1}}
\usepackage{multirow}

\floatstyle{ruled}

\usepackage[symbol]{footmisc}

\newcommand{\ie}{\textit{i}.\textit{e}. }
\newcommand{\eg}{\textit{e}.\textit{g}. }

\begin{document}
\pagestyle{headings}
\mainmatter
\def\ECCVSubNumber{2191}  

\title{Improved Mutual Mean-Teaching for Unsupervised Domain Adaptive Re-ID} 

\titlerunning{Improved Mutual Mean-Teaching}
%
\author{Yixiao Ge
\and
Shijie Yu
\and 
Dapeng Chen}
\authorrunning{Y. Ge et al.}
%
\institute{
The Chinese University of Hong Kong \\
\email{yxge@link.cuhk.edu.hk}
}
\maketitle

\begin{abstract}
In this technical report, we present our submission to the VisDA Challenge in ECCV 2020 and we achieved one of the top-performing results on the leaderboard.
Our solution is based on Structured Domain Adaptation (SDA) \cite{ge2020structured} and Mutual Mean-Teaching (MMT) \cite{ge2020mutual} frameworks. 
SDA, a domain-translation-based framework, focuses on carefully translating the source-domain images to the target domain.
MMT, a pseudo-label-based framework, focuses on conducting pseudo label refinery with robust soft labels.
Specifically, there are three main steps in our training pipeline.
(i) We adopt SDA to generate source-to-target translated images, and (ii) such images serve as informative training samples to pre-train the network. (iii) The pre-trained network is further fine-tuned by MMT on the target domain. Note that we design an improved MMT (dubbed MMT+) to further mitigate the label noise by modeling inter-sample relations across two domains and maintaining the instance discrimination.
Our proposed method achieved 74.78\% accuracies in terms of mAP, ranked the 2nd place out of 153 teams.
\footnote[2]{Code of this work is available at \url{https://github.com/yxgeee/VisDA-ECCV20}. Video introduction is available at \url{https://youtu.be/Ox-ZJhgFwSU} or \url{https://www.bilibili.com/video/BV14V411U7mb}.}
\end{abstract}

\section{Introduction}

\begin{figure}[t]
\centering
\includegraphics[width=1.0\linewidth]{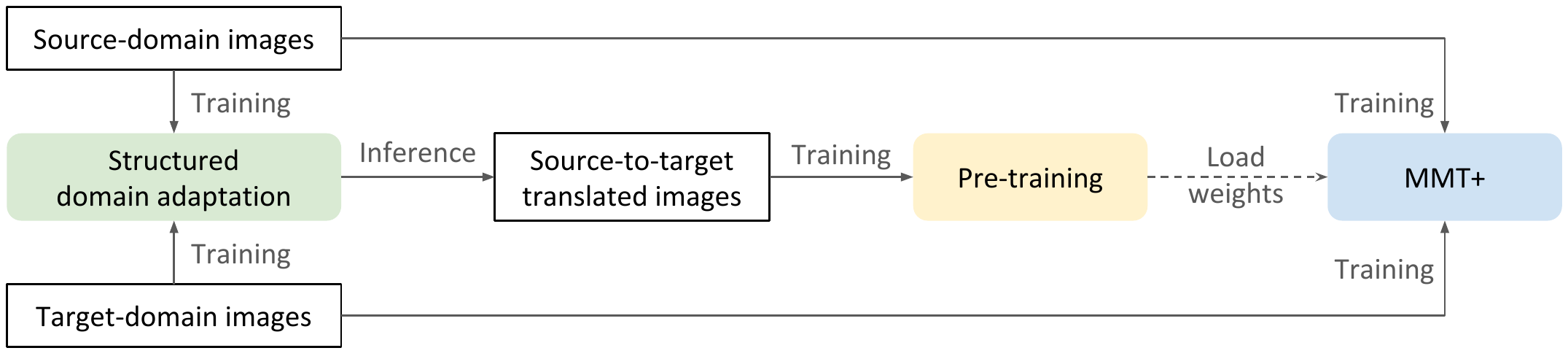}
\caption{The training pipeline of our proposed method, which consists of three steps: structured domain adaptation (SDA), pre-training with source-to-target translated images and fine-tuning on the target domain with the improved MMT framework.}
\label{fig:intro}
\end{figure}

Accurate person re-identification (re-ID) is at the core of smart city systems, which aims at retrieving the same person's images across multiple cameras. Although deep learning-based methods have achieved satisfying re-ID performances by training with large-scale datasets, inevitable domain gaps between different camera systems prevent the trained networks from being directly deployed on a new scene. 
Unsupervised domain adaptation (UDA) is therefore proposed to transfer the learned knowledge from the labeled source domain (dataset) to the unlabeled target domain (dataset).

In ECCV 2020, the VisDA Challenge introduced a synthetic$\to$real UDA task on person re-ID. 
The synthetic source-domain data generated by the Unity engine \cite{riccitiello2015john} have the same style as PersonX \cite{sun2019dissecting} and their labels can be naturally provided. 
The unlabeled target-domain images are collected from real-world scenarios.
It is a much challenging problem due to the fact that 1) there are larger domain gaps between synthetic and real scenarios than those between real and real scenarios and 2) the distribution of unlabeled target-domain data is much more realistic than previous benchmarks.
Specifically, existing benchmarks for UDA re-ID (\eg Market-1501 \cite{market}$\to$DukeMTMC-reID \cite{dukemtmc}) generally construct the target-domain dataset by simply removing the labels from public datasets, however, such public datasets have already been manually organized, which could not well match the distribution of unlabeled datasets in practical use.
In contrast, VisDA provides a noisy target-domain dataset with more practical settings.

Existing methods tackling the problem of UDA re-ID can be divided into two main categories, domain-translation-based methods \cite{deng2018image,wei2018person,chen2019instance,ge2020structured} and pseudo-label-based methods \cite{song2018unsupervised,yang2019selfsimilarity,zhang2019self,ge2020mutual,zhai2020ad,zhong2019invariance,yu2019unsupervised,wang2020unsupervised}. 
Domain-translation-based methods target at translating the source-domain images to have the target-domain style while well preserving their original IDs \cite{deng2018image,wei2018person} or inter-sample relations \cite{ge2020structured}. Such kind of methods provides a plausible way to make use of source-domain images and their valuable ground-truth identities.
Pseudo-label-based methods aim at learning the distribution of unlabeled target-domain data with pseudo labels, where the pseudo labels are generated by either clustering instance features \cite{song2018unsupervised,yang2019selfsimilarity,zhang2019self,ge2020mutual,zhai2020ad} or measuring similarities with exemplar features \cite{zhong2019invariance,yu2019unsupervised,wang2020unsupervised}. 
Although pseudo-label-based methods could achieve superior performance than domain-translation-based methods, we argue that they are complementary to each other and can work together to achieve optimal performance.

To effectively take advantage of both domain-translation-based and pseudo-label-based methods, we introduce a training pipeline with Structured Domain Adaptation (SDA) \cite{ge2020structured} and Mutual Mean-Teaching (MMT) \cite{ge2020mutual} frameworks. 
SDA, one of the state-of-the-art domain-translation-based methods \cite{deng2018image,wei2018person,chen2019instance,ge2020structured},
adopts CycleGAN \cite{zhu2017unpaired} architecture to perform image-to-image translation. An online relation-consistency regularization is introduced to maintain inter-sample relations during the training of SDA.
Domain translation is crucial in this task due to the evident domain gaps between synthetic data and real-world data.

MMT, one of the state-of-the-art pseudo-label-based methods \cite{song2018unsupervised,yang2019selfsimilarity,zhang2019self,ge2020mutual,zhai2020ad,zhong2019invariance,yu2019unsupervised,wang2020unsupervised}, proposes to conduct pseudo label refinery with reliable soft labels, which are generated online in a mutual teaching pipeline.
We further improve the MMT framework, namely MMT+, by jointly training with both source-domain images and target-domain images to model complex inter-sample relations across two domains.
To further mitigate the effects caused by noisy pseudo labels, we add the MoCo \cite{he2019momentum} loss to maintain the instance discrimination.
Since the mean networks in the MMT framework are similar to the momentum encoders in MoCo \cite{he2019momentum}, the MoCo loss can be easily utilized without extra costs.

As shown in Fig. \ref{fig:intro}, there are three main training stages in our introduced pipeline.
(i) An SDA framework is trained to translate source-domain images to the target domain.
(ii) Source-to-target translated images serve as training samples to pre-train the network with ground-truth identities. The network can then be roughly adapted to the target domain.
(iii) The pre-trained network is further fine-tuned on the target domain with the proposed MMT+ framework. Both labeled source-domain raw images and unlabeled target-domain images are used for training.

The contributions of this work could be summarized as three-fold. 
(1) We introduce a joint pipeline to properly make use of both the domain-translation-based and pseudo-label-based frameworks, which are complementary to each other.
(2) We propose to improve the state-of-the-art MMT \cite{ge2020mutual} framework by modeling the inter-/intra-identity relations across two domains and preserving the instance discrimination to mitigate the effects caused by noisy pseudo labels.
(3) Our proposed method achieved one of the top-performing results on the leaderboard, yielding 74.78\% accuracies in terms of mAP, which ranked the 2nd place in the VisDA Challenge.

%
%
%

\section{Preliminary}

\subsection{Source-domain Pre-training}
\label{sec:source_pretrain}


Given source-domain images $\sX^s$, the neural network $\mathcal{F}^s$ is trained to transform each sample $x^s \in \sX^s$ into a feature vector $\vf^s=\mathcal{F}^s(x^s)$, which could be used to predict its ground-truth identity $y^s$ with a learnable classifier $\mathcal{C}^s: \vf^s \to \{1,\cdots,p^s \}$, where $p^s$ is the number of identities in the source domain. 
A classification loss in the form of cross-entropy loss $\ell_\text{ce}$ and a softmax-triplet loss in the form of binary cross-entropy loss $\ell_\text{bce}$ are adopted jointly for training,
{
\begin{align} 
\label{eq:source_pretrain} 
&\mathcal{L}^s_\text{cls}(\mathcal{F}^s,\mathcal{C}^s) = \mathbb{E}_{x^s \sim \sX^s}\left[ \ell_\text{ce} ( \mathcal{C}^s(\vf^s), y^s  )\right], \\
&\mathcal{L}^s_\text{tri}(\mathcal{F}^s) = 
\mathbb{E}_{x^s \sim \sX^s}\left[\ell_\text{bce}(\mathcal{T}(\vf^s),\1) \right],
\end{align}}%
where
\begin{align}
\mathcal{T}(\vf^s) = \frac{\exp(\| \vf^s-\vf^s_n\|)}{\exp(\| \vf^s-\vf^s_p\|)+\exp(\| \vf^s-\vf^s_n\|)},
\end{align}
and the subscripts $_p, _n$ denote the mini-batch's hardest positive and negative feature indexes for the anchor $\vf^s$. The overall loss function for source-domain pre-training is $\mathcal{L}^s=\mathcal{L}^s_\text{cls}+\mathcal{L}^s_\text{tri}$.

\subsection{Clustering-based Baseline Training}
\label{sec:cluster_baseline}

Clustering-based pipeline serves as a strong baseline model for pseudo-label-based methods, which alternates between (i) generating pseudo classes by clustering target-domain instances $\sX^t$ and (ii)  training the network $\mathcal{F}^t$ with generated pseudo classes. 
Specifically, after loading the pre-trained weights of $\mathcal{F}^s$ to $\mathcal{F}^t$, $\sX^t$'s encoded features $\{\vf^t\}$ are clustered into $\hat{p}^t$ classes and images within the same cluster are assigned the same label.
Generally, the density-based clustering algorithm (\eg DBSCAN) is adopted as we do not know the number of the target-domain identities.
Note that $\hat{p}^t$ is automatically measured by density-based clustering.
Similar to source-domain pre-training, a classification loss and a softmax-triplet loss are adopted,
{
\begin{align} 
\label{eq:cluster_baseline} 
&\mathcal{L}^t_\text{cls}(\mathcal{F}^t,\mathcal{C}^t) = \mathbb{E}_{x^t \sim \sX^t}\left[ \ell_\text{ce} ( \mathcal{C}^t(\vf^t), \hat{y}^t  )\right], \\
&\mathcal{L}^t_\text{tri}(\mathcal{F}^t) = 
\mathbb{E}_{x^t \sim \sX^t}\left[\ell_\text{bce}(\mathcal{T}(\vf^t),\1) \right],
\end{align}}%
where $\hat{y}^t$ denotes the pseudo label for unlabeled data $x^t$. The overall loss function for clustering-based baseline training is $\mathcal{L}^t=\mathcal{L}^t_\text{cls}+\mathcal{L}^t_\text{tri}$. The generated pseudo labels are updated before each epoch.

\section{Proposed Approach}

We propose to properly make use of both the domain-translation-based method (\ie SDA) and the pseudo-label-based method (\ie MMT) with a three-stage training pipeline: 
(i) training a structured domain adaptation framework to carefully translate source-domain images to the target domain;
(ii) the translated images serve as informative training samples to pre-train the network and roughly adapt the model onto the target domain;
(iii) the pre-trained network is further fine-tuned on the target domain with an improved mutual mean-teaching framework.

\subsection{Structured Domain Adaptation}
\begin{figure}[t]
\centering
\includegraphics[width=1.0\linewidth]{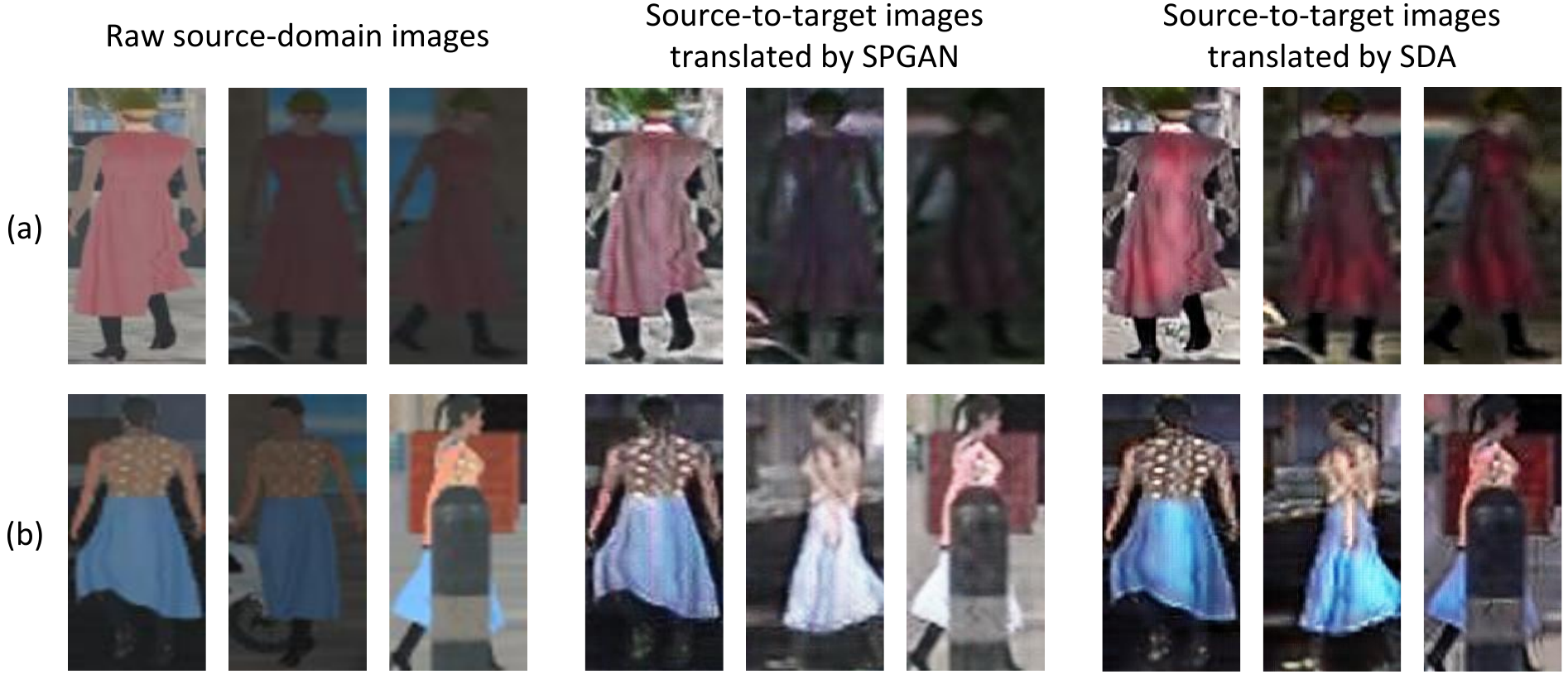}
\caption{Source-to-target translated images. Compared to SPGAN \cite{deng2018image}, the SDA \cite{ge2020structured} could better preserve inter-sample relations, \eg maintaining consistent appearance for person images of the same identity.}
\label{fig:sda}
\end{figure}
Domain translation is crucial in the synthetic$\to$real adaptation task since the gaps between source and target domains are significant. Directly pre-training the network with raw source-domain data as described in Sec. \ref{sec:source_pretrain} could only achieve limited performance on the target domain, which will result in inaccurate pseudo labels. 
We adopt the state-of-the-art domain-translation-based framework, Structured Domain Adaptation (SDA) \cite{ge2020structured}, to translate source-domain images to have the target-domain style.
SDA adopts CycleGAN \cite{zhu2017unpaired} architecture as the backbone, and 
introduces an online relation-consistency regularization to maintain the inter-sample relations instead of simple IDs. 
Specifically, the inter-sample relations in SDA are measured online by source-domain and target-domain encoders.
We use the pre-trained $\mathcal{F}^s$ (Sec. \ref{sec:source_pretrain}) as the source-domain encoder and $\mathcal{F}^t$ trained by clustering-based baseline (Sec. \ref{sec:cluster_baseline}) as the target-domain encoder.
The relation-consistency regularization can therefore be formulated as a soft binary cross-entropy loss
\begin{align}
\mathcal{L}_\text{rc}(\mathcal{G}^{s\to t}) = 
\mathbb{E}_{x^s \sim \sX^s}\left[\ell_\text{bce}(\mathcal{T}(\vf^{s\to t}), \mathcal{T}(\vf^s)) \right],
\end{align}
where $\mathcal{G}^{s\to t}$ is the source-to-target generator, $\vf^{s\to t}=\mathcal{F}^t(x^{s\to t})$ and $\vf^s=\mathcal{F}^s(x^s)$.
Besides $\mathcal{L}_\text{rc}$, conventional cycle generation losses used by CycleGAN are needed.
As illustrated in Fig. \ref{fig:sda}, SDA can generate informative training samples, and the generated samples could be trained to achieve better pre-training performance.

\subsection{Pre-training with Source-to-target Translated Images}
Given the trained $\mathcal{G}^{s\to t}$ in SDA, we could translate the source-domain images $\sX^s$ to the target domain, denoted as $\sX^{s\to t}$. 
By adopting the similar classification loss and softmax-triplet loss in Sec. \ref{sec:source_pretrain}, we re-train the network $\mathcal{F}^s$ with $\sX^{s\to t}$ and ground-truth labels $\sY^s$
{
\begin{align} 
\label{eq:sda_pretrain} 
&\mathcal{L}^s_\text{cls}(\mathcal{F}^s,\mathcal{C}^s) = \mathbb{E}_{x^{s\to t} \sim \sX^{s\to t}}\left[ \ell_\text{ce} ( \mathcal{C}^s(\hat{\vf}^{s\to t}), y^s  )\right], \\
&\mathcal{L}^s_\text{tri}(\mathcal{F}^s) = 
\mathbb{E}_{x^{s\to t} \sim \sX^{s\to t}}\left[\ell_\text{bce}(\mathcal{T}(\hat{\vf}^{s\to t}),\1) \right],
\end{align}}%
where $\hat{\vf}^{s\to t}=\mathcal{F}^s(x^{s\to t})$.
By pre-training with translated images $\sX^{s\to t}$, the network $\mathcal{F}^s$ can be roughly adapted to the target domain with much better performance than $\mathcal{F}^s$ pre-trained with $\sX^{s}$ in Sec. \ref{sec:source_pretrain}.

\subsection{Improved Mutual Mean-Teaching}
\begin{figure}[t]
\centering
\includegraphics[width=1.0\linewidth]{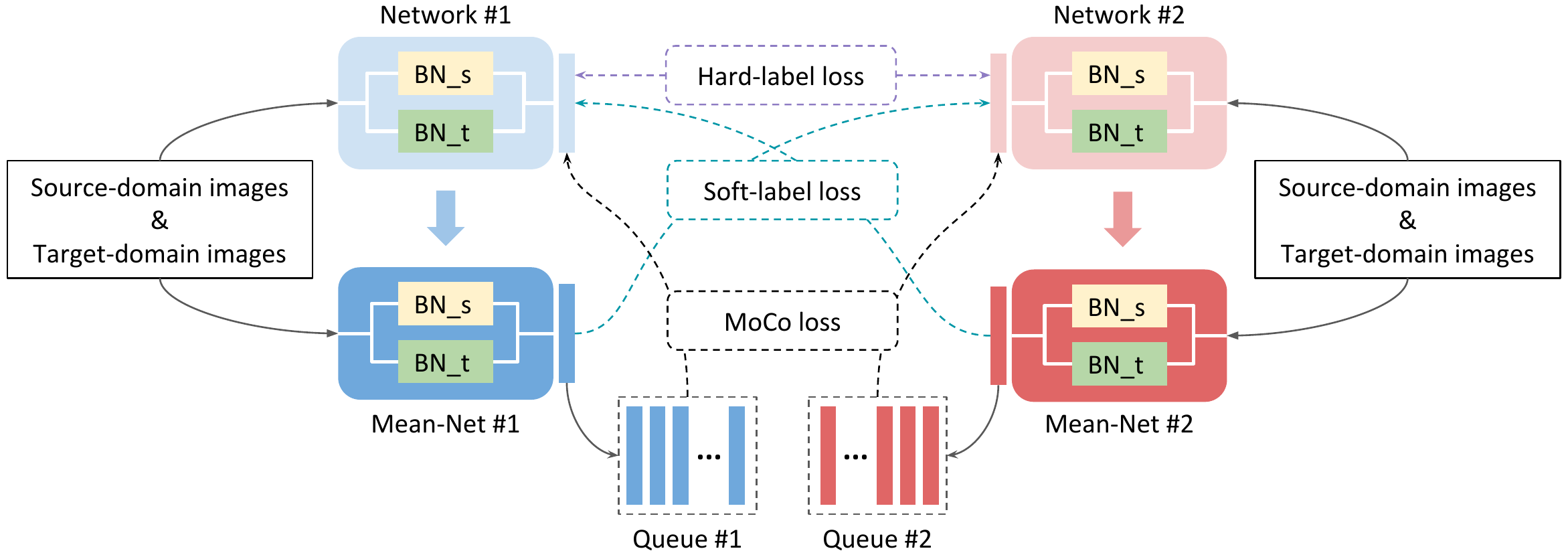}
\caption{We improve the Mutual Mean-Teaching (MMT) \cite{ge2020mutual} framework by jointly training with both source-domain and target-domain images to model complex inter-sample relations, and adding an instance discriminative MoCo \cite{he2019momentum} loss to further mitigate the effects caused by noisy pseudo labels.}
\label{fig:mmt+}
\end{figure}
The Mutual Mean-Teaching (MMT) \cite{ge2020mutual} framework adopts a couple of networks, denoted as $\mathcal{F}^t_1$ and $\mathcal{F}^t_2$, and each network's mean-teacher network supervises the training of the other network by predicting robust soft labels. 
The original MMT framework is only trained with target-domain data and generated pseudo labels, however, we argue that properly modeling the inter-samples relations across two domains is critical to the final performance.
We introduce an improved mutual mean-teaching framework (MMT+) by jointly training with two domains' raw images and minimize the domain gaps in mini-batches with domain-specific BatchNorms \cite{Chang_2019_CVPR} (Fig. \ref{fig:mmt+}).
Specifically, we denote the joint dataset as $\sX=\sX^s \cup \sX^t$. 
The cross-domain inter-sample relations are modeled by constructing a joint label system with both source-domain ground-truth IDs and target-domain pseudo IDs.
Given an encoded feature $\vf=\mathcal{F}^t(x)$, it is trained to predict its own label among all the identities across two domains.

Let's denote the mean networks as $\E[\mathcal{F}^t_1]$ and $\E[\mathcal{F}^t_2]$, 
logits predicted by one mean network serve as soft targets to train the other network by a soft cross-entropy loss
\begin{align}
\label{eq:mmt}
\mathcal{L}^t_\text{soft}(\mathcal{F}^t_1, \mathcal{F}^t_2, \mathcal{C}^t_1, \mathcal{C}^t_2)=&-\mathbb{E}_{x \sim \sX}\Big[ \E[\mathcal{C}^t_2]( \E[\mathcal{F}^t_2](x))\cdot \log \mathcal{C}^t_1(\mathcal{F}^t_1(x)) \nonumber\\
&+ \E[\mathcal{C}^t_1](\E[\mathcal{F}^t_1](x))\cdot \log \mathcal{C}^t_2(\mathcal{F}^t_2(x)) \Big],
\end{align}
where the logits need to be normalized by a softmax operation.
Instead of conventional back-propagation,
the weights of mean networks are updated with a moving average formulation $\E[\theta]=\alpha\E[\theta]+(1-\alpha)\theta$, where $\alpha=0.999$ is the momentum hyper-parameter.
Besides the soft-label loss in Eq. (\ref{eq:mmt}), hard-label loss similar to Eq. (\ref{eq:cluster_baseline}) is adopted
{
\begin{align} 
&\mathcal{L}^t_\text{hard}(\mathcal{F}^t_1, \mathcal{F}^t_2, \mathcal{C}^t_1, \mathcal{C}^t_2) = \mathbb{E}_{x \sim \sX}\left[ \ell_\text{ce} ( \mathcal{C}^t_1(\mathcal{F}^t_1(x)), y ) + \ell_\text{ce} ( \mathcal{C}^t_2(\mathcal{F}^t_2(x)), y )\right]. 
\end{align}}%

In the experiments, we find that the pseudo labels are much noisy due to the noisy distribution of identities in the target domain.
In order to further mitigate the effects caused by pseudo label noise, we propose to adopt a MoCo \cite{he2019momentum} loss to maintain the instance discrimination.
Specifically, the MoCo loss needs a queue with a fixed length to cache the features encoded by the momentum encoder, which is almost the same as our mean network. So we save the features encoded by our mean networks in the queue, and such features act as negative samples in the MoCo loss. 
Due to the coupled networks in MMT, we need two queues to serve individually for each network.
The overall MoCo loss is formulated as a contrastive loss
\begin{align}
\mathcal{L}^t_\text{moco}(\mathcal{F}^t_1, &\mathcal{F}^t_2)=-\mathbb{E}_{x \sim \sX}\Bigg[ \log \frac{\exp (\langle \mathcal{F}^t_1(x),\E[\mathcal{F}^t_1](x) \rangle/\tau)}{\exp (\langle \mathcal{F}^t_1(x),\E[\mathcal{F}^t_1](x) \rangle/\tau)+\sum_{\vk^-_1} \exp (\langle \mathcal{F}^t_1(x),\vk^-_1 \rangle/\tau)}  \nonumber\\
&+ \log \frac{\exp (\langle \mathcal{F}^t_2(x),\E[\mathcal{F}^t_2](x) \rangle/\tau)}{\exp (\langle \mathcal{F}^t_2(x),\E[\mathcal{F}^t_2](x) \rangle/\tau)+\sum_{\vk^-_2} \exp (\langle \mathcal{F}^t_2(x),\vk^-_2 \rangle/\tau)}  \Bigg],
\end{align}
where $k^-$ denotes the features from queue and $\tau=0.7$ is the temperature hyper-parameter. The overall loss for the proposed MMT+ is
\begin{align}
\mathcal{L}^t_\text{mmt+}=\lambda_\text{soft}\mathcal{L}^t_\text{soft}+(1-\lambda_\text{soft})\mathcal{L}^t_\text{hard}+\lambda_\text{moco}\mathcal{L}^t_\text{moco},
\end{align}
where $\lambda_\text{soft}=0.5$, $\lambda_\text{moco}=0.1$ are the weighting parameters. 

\section{Experiments}

\subsection{Dataset and Evaluation Metric}

There are four subsets provided by the VisDA Challenge\footnote[3]{\url{https://github.com/Simon4Yan/VisDA2020}}: source\_train, target\_train, target\_val and target\_test, where only source\_train and target\_train can be used for training.
The synthetic source-domain dataset is generated based on Unity \cite{riccitiello2015john}.
Specifically, the source\_train set consists of 20,280 images out of 700 identities shot from 6 cameras in total.
The target-domain dataset consists of real-world images captured from 5 cameras, \ie 13,198 images for training, 377 images for the query of target\_val, 3,600 images for the gallery of target\_val, 1,578 images for the query of target\_test and 24,006 images for the gallery of target\_test.
Mean Average Precision (mAP) and Cumulated Matching Characteristics (CMC) accuracies are adopted to test the methods’ performances, where only the top-$100$ matches are considered for evaluation.

\subsection{Implementation Details}
\label{sec:imp}

We implement our framework in PyTorch \cite{pytorch} and all the person images are resized to 384$\times$128.
We adopt ImageNet \cite{deng2009imagenet}-pretrained networks up to the global average pooling layer as the backbone, where GeM pooling \cite{radenovic2018fine} is adopted to replace the global average pooling layer for optimal performance. 
Adam optimizer is adopted with a weight decay of $0.0005$.

\subsubsection{Stage I: Structured domain adaptation.}
We adopt 8 GPUs for training the SDA, where each mini-batch contains 32 source-domain images of 8 ground-truth classes and 32 randomly sampled target-domain images. The training scheme iterates for 100 epochs and there are 200 iterations in each epoch, where the learning rate ($\text{lr}=0.0002$) is constant for the first 50 epochs and then gradually decreases to 0 for another 50 epochs following the formula $ \text{lr} = \text{lr} \times (1.0-\max (0, \text{epoch}-50)/50) $.

\subsubsection{Stage II: Pre-training with source-to-target translated images.}
We adopt 4 GPUs for pre-training, where each mini-batch contains 64 source-to-target images of 16 ground-truth classes. Auto augmentation \cite{cubuk2018autoaugment} is adopted for pre-training. The initial learning rate is set to $0.00035$ and is decreased to 1/10 of its previous value on the 40th and 70th epoch in the total 120 epochs. Each epoch has 200 iterations.

\subsubsection{Stage III: Improved mutual mean-teaching.}
We adopt 4 GPUs for MMT+ training, where each mini-batch contains 64 source-domain images of 16 ground-truth classes and 64 target-domain images of 16 pseudo classes. Random erasing \cite{zhong2017random} is adopted. 
Domain-specific BNs \cite{Chang_2019_CVPR} are adopted in this stage to minimize the domain gaps in each mini-batch. The learning rate is fixed to $0.00035$ for overall 50 training epochs, where each epoch has 200 iterations. 
We use DBSCAN \cite{ester1996density} and Jaccard distance \cite{zhong2017re} with k-reciprocal nearest neighbors for clustering before each epoch, where $k=20$. 
As for DBSCAN, the maximum distance between neighbors is set as 0.6 and the minimal number of neighbors for a dense point is set as 4. 
The length of queue in MoCo loss is set as 12,800.
Arcface loss or Cosface loss is adopted in this step to replace the simple classification loss.

\subsubsection{Post-processing.}
(1) We adopt ResNet50-IBN \cite{pan2018two} as the backbone for the training stage I to generate source-to-target training samples. ResNeSt50 \cite{zhang2020resnest}, ResNeSt101 \cite{zhang2020resnest}, DenseNet169-IBN \cite{huang2017densely,pan2018two} and ResNeXt101-IBN \cite{Xie2016,pan2018two} are adopted for ensembling after training in stage II and III. Specifically, features encoded by the above backbones are concatenated and $L_2$-normalized.
(2) Following \cite{zhu2020voc}, we train a camera classification network to predict the camera similarities between testing images. Let's denote the camera network as $\mathcal{F}_c$, then the image similarity between a query image $q$ and a gallery image $k$ is $s(q,k)=\|\vf_q-\vf_k\|-0.1 \|\mathcal{F}_c (q)-\mathcal{F}_c (k) \|$. $\mathcal{F}_c$ adopts ResNeSt50 \cite{zhang2020resnest} as the backbone.
(3) We adopt the re-ranking technique \cite{zhong2017re} with $k_1=30, k_2=6, \lambda=0.3$.

\subsection{Quantitative Results}

\begin{table}[th]
	\caption{Competition results of the VisDA Challenge in ECCV 2020. The results are evaluated on the target\_test set. Our result is in \textbf{bold}.}
	\label{tab:sota}
	\begin{center}
	\begin{tabular}{P{4cm}|C{2cm}C{2cm}}
	\hline
	Team Name & mAP(\%) & top-1(\%) \\ 
	\hline \hline
	Vimar Team & 76.56 &	84.25 \\
	\textbf{Ours} & \textbf{74.78} &	\textbf{82.86} \\
	Xiangyu & 72.39 & 83.85 \\
	\hline
	\end{tabular}
	\end{center}
\end{table}

\subsubsection{Comparison with other teams.}
As shown in Tab. \ref{tab:sota}, the introduced method achieved the 2nd place in terms of mAP accuracy, \ie 74.78\%.

\begin{table}[th]
	\caption{Ablation study on the effectiveness of source-to-target images translated by SDA. The results are evaluated on the target\_val set. The pre-training for this ablation study adopts the backbone of ResNet50-IBN \cite{pan2018two}.}
	\label{tab:sda}
	\begin{center}
	\begin{tabular}{P{8cm}|C{2cm}C{2cm}}
	\hline
	Images for pre-training & mAP(\%) & top-1(\%) \\ 
	\hline \hline
	Raw source-domain images & 61.0 & 71.6 \\
	Source-to-target images translated by SPGAN \cite{deng2018image}\footnote{Downloaded from \url{https://github.com/Simon4Yan/VisDA2020}.} & 68.2 & 75.1 \\
	\textbf{Source-to-target images translated by SDA \cite{ge2020structured}} & \textbf{71.2} &	\textbf{79.3} \\
	\hline
	\end{tabular}
	\end{center}
\end{table}

\subsubsection{Effectiveness of SDA.}
To verify the effectiveness of source-to-target images translated by SDA for pre-training,
we compared the pre-training performance with raw images and SPGAN translated images in Tab. \ref{tab:sda}.

\begin{table}[th]
	\caption{Ablation study on the effectiveness of the improved MMT. The results are evaluated on the target\_val set. The experiments in this ablation study adopts the backbone of ResNet50-IBN \cite{pan2018two}. All the post-processing techniques except the ensembling as described in Sec. \ref{sec:imp} are used.}
	\label{tab:mmt}
	\begin{center}
	\begin{tabular}{P{6cm}|C{2cm}C{2cm}}
	\hline
	Pseudo-label-based method & mAP(\%) & top-1(\%) \\ 
	\hline \hline
	Original MMT \cite{ge2020mutual} & 78.4 & 86.5 \\
	\textbf{Our MMT+} & \textbf{81.2} &	\textbf{87.3} \\
	\hline
	\end{tabular}
	\end{center}
\end{table}

\subsubsection{Effectiveness of improved MMT.}
Compared to the original MMT, our proposed MMT+ achieves 2.6\% improvements in terms of mAP (Tab. \ref{tab:mmt}).

\begin{table}[th]
	\caption{Performance of different backbones for MMT+. The results are evaluated on the target\_val set. All the post-processing techniques as described in Sec. \ref{sec:imp} are used.}
	\label{tab:backbone}
	\begin{center}
	\begin{tabular}{P{6cm}|C{2cm}C{2cm}}
	\hline
	Backbone for MMT+ & mAP(\%) & top-1(\%) \\ 
	\hline \hline
	ResNeSt50 \cite{zhang2020resnest} & 83.6 & 89.4 \\
	ResNeSt101 \cite{zhang2020resnest} & 82.7 & 89.1 \\
	DenseNet169-IBN \cite{huang2017densely,pan2018two} & {84.1} & {90.5} \\
	ResNeXt101-IBN \cite{Xie2016,pan2018two} & 83.5 & 88.9 \\
	Ensemble & \textbf{86.3} & \textbf{91.2} \\
	\hline
	\end{tabular}
	\end{center}
\end{table}

\subsubsection{Performance of different backbones for MMT+.}
As shown in Tab. \ref{tab:backbone}, we report the performances when training with different backbones. ResNeSt50 \cite{zhang2020resnest}, ResNeSt101 \cite{zhang2020resnest}, DenseNet169-IBN \cite{huang2017densely,pan2018two} and ResNeXt101-IBN \cite{Xie2016,pan2018two} are adopted for ensembling in the final submission.

\section{Conclusion and Discussion}

In this work, we properly make use of both domain-translation-based and pseudo-label-based frameworks, which are both important for the unsupervised domain adaptation task on person re-ID. We also improve the state-of-the-art MMT framework to achieve better performance. Although we did not achieve the 1st place in the VisDA Challenge this year, we believe that the introduced methods have great potential in this research field. Also, there leaves some room for improvement and we will keep trying to make more progress.

\clearpage
%
%
\bibliographystyle{splncs04}
\bibliography{visda}
\end{document}

%% file: math_commands.tex
\usepackage{amsmath,amsfonts,bm}

\def\eqref#1{equation~\ref{#1}}

\def\1{\bm{1}}

\newcommand{\test}{\mathcal{D_{\mathrm{test}}}}

\def\vf{{\bm{f}}}

\def\vk{{\bm{k}}}

\DeclareMathAlphabet{\mathsfit}{\encodingdefault}{\sfdefault}{m}{sl}
\SetMathAlphabet{\mathsfit}{bold}{\encodingdefault}{\sfdefault}{bx}{n}

\def\sX{{\mathbb{X}}}
\def\sY{{\mathbb{Y}}}

\newcommand{\E}{\mathbb{E}}

